\documentclass[11pt,a4paper]{article}
\usepackage[hyperref]{emnlp2020}
\usepackage{times}
\usepackage{latexsym}
\usepackage{amsfonts}
\usepackage{amsmath}
\usepackage{graphicx}
\usepackage{booktabs}
\usepackage[ruled,vlined]{algorithm2e}
\usepackage{bbm}

\usepackage{graphicx,subcaption}

\usepackage{microtype}

\aclfinalcopy 

\usepackage{todonotes}

\newcommand{\KLD}[2]{D_{\mathrm{KL}} \left(  #1 \middle\| #2 \right) }

\DeclareMathOperator*{\gargmax}{greedy\,argmax}

\title{Preventing Posterior Collapse with Levenshtein Variational Autoencoder}
\author{Serhii Havrylov \\
  ILCC, University of Edinburgh \\
  \texttt{s.havrylov@ed.ac.uk} \\\And
  Ivan Titov \\
  ILCC, University of Edinburgh \\
  ILLC, University of Amsterdam \\
  \texttt{ititov@inf.ed.ac.uk} \\}

\date{}

\begin{document}
\maketitle

\begin{abstract}
Variational autoencoders (VAEs) are a standard framework for inducing latent variable models that have been shown effective in learning text representations as well as in text generation.
The key challenge with using  VAEs is the {\it posterior collapse} problem: learning tends to converge to trivial solutions where the generators ignore latent variables.
In our Levenstein VAE, we propose to replace the evidence lower bound (ELBO) with a new objective which is simple to optimize and prevents  posterior collapse.
Intuitively, it corresponds to generating a sequence from the autoencoder and encouraging the model to predict an optimal continuation according to the Levenshtein distance (LD) with the reference sentence at each time step in the generated sequence.
We motivate the method from the probabilistic perspective by showing that it is closely related to optimizing a bound on the intractable Kullback-Leibler divergence of an LD-based kernel density estimator from the model distribution.
With this objective, any generator disregarding latent variables will incur large penalties and hence posterior collapse does not happen.
We relate our approach to policy distillation~\cite{RossGB11} and dynamic oracles~\cite{GoldbergN12}.
By considering Yelp and SNLI benchmarks, we show that Levenstein VAE produces more informative latent representations than alternative approaches to preventing posterior collapse.

\end{abstract}

\section{Introduction}
A latent variable model is a statistical model that assumes that a sequence \(x\) is generated by some random process, involving an unobserved random variable \(z\).
While latent variable models have a long history  in natural language processing (e.g., \citet{brown1993mathematics,blei2003latent}), recently, much work has focused on {\it deep generative modeling}, i.e. combining neural networks and latent variable modeling.
This direction is attractive as it brings together advantages of neural networks (i.e., providing powerful function approximators) with the benefits of latent variable modeling (e.g., natural ways of specifying inductive biases or supporting semi-supervised learning).
Deep generative models have been shown effective in producing accurate language models~\cite{abs-1904-08194}, as well as, in inducing informative representations for downstream tasks~\cite{FuLLGCC19}. 
Typically, these methods rely on the variational autoencoding (VAE) framework~\citep{KingmaW13, RezendeMW14}.
The variational autoencoders optimize the lower bound (ELBO) on the (usually intractable) marginal likelihood \(p_\theta(x)\).
One of the key issues with using variational autoencoders is the {\it posterior collapse} problem~\cite{BowmanVVDJB16}: learning often converges to a trivial optimum, a solution where the generator ignores the latent variable $z$ when generating $x$.
While a number of approaches have been proposed to address this issue (see section \ref{sec:related_work}), they typically modify the architecture of a generator, are much slower in comparison to the vanilla version or are problematic to use with high-dimensional latent variables.

In our work, we propose to replace ELBO with a new objective which is easy to optimize and which prevents the posterior collapse. 
We motivate the objective by considering a kernel density estimator (KDE) that relies on the Levenshtein distance (LD) as its kernel. 
We derive our objective by considering the maximization of the upper bound on the intractable Kullback-Leibler divergence of   the  kernel density estimator from the model distribution $p_\theta(x)$.
An important property of this objective is that any generator ignoring latent space will incur a large penalty, and, as a result, posterior collapse does not occur.
This contrasts with the usual maximum likelihood objective used in VAEs which a strong (e.g., recurrent) autoregressive model can fit well without relying on any latent information.

Intuitively, in our {\it Levenshtein VAE}, the reconstruction term of the vanilla VAE gets replaced with a new loss term.
This term corresponds to generating a sequence from the autoencoder and forcing the model to predict an optimal continuation at each time step in the generated sequence.
The set of optimal continuations is computed with respect to the target sequence using the Levenshtein distance (see an example in Figure~\ref{tab:ocp}).
Note that these optimal continuations can be effectively computed with dynamic programming ~\citep{SabourCN19} and are closely related to dynamic oracles~\cite{GoldbergN12}.

We consider Yelp and SNLI datasets and compare representations learned with Levenshtein VAE against the vanilla VAE and popular alternative approaches to tackling posterior collapse, $\beta$-VAE~\cite{HigginsMPBGBML17} and cyclical annealing VAE~\cite{FuLLGCC19}.
We 
observe that representations produced by our approach, are considerably more informative than those of the alternatives, as measured using diagnostic classifiers.
We also consider language modeling and measure perplexity. 
Note that the perplexity-based evaluation examines the model predictions when given
true (`gold-standard') prefixes.
This matches the maximum likelihood (and hence ELBO) objectives but does not directly correspond to the way Levenshtein VAE has been optimized.
Nevertheless, when simply interpolated with the objective of standard VAE, our model both does not suffer from the posterior collapse and achieves strong results on this benchmark.
 
While we experiment with a specific (though standard) architecture and consider non-conditional generation, our approach is general.
It can, in principle, be applied in most applications where VAEs have been shown effective~\citep{ShenLBJ17, HuYLSX17}.
Moreover, Levenshtein distance can be replaced with other alternatives (e.g., its generalization to ngram similarities~\cite{kondrak2005n}), as long as optimal continuations are efficiently computable. 
Our main contributions can be summarized as follows
\begin{itemize}
 \item we introduce a new and simple approach to preventing posterior collapse in variational autoencoders;
 \item we motivate it from the probabilistic perspective, showing that it is a valid loss for training a latent variable model;
 \item we demonstrate its effectiveness on representation learning and language modeling benchmarks.
\end{itemize}

\section{Preliminaries}


\subsection{Latent variable models for sequence generation}
 

When using a latent variable model, 
 the probability of a sequence \(x\) is defined as: \(p_\theta(x) = \mathbb{E}_{p(z)}\left[\prod_{t=1}^{|x|}p_\theta(x_t|x_{<t}, z)\right]\).
Usually, parameters of such model are estimated by optimizing the marginal log likelihood of the data: \(\sum_{x\in\mathcal{D}}\log{p_\theta(x)}\).
Directly minimising this objective function is problematic due to the absence of the closed-form solution for the integral, which is caused by RNN parameterisation of each conditional distribution.
Even though na\"ive application of Monte Carlo (MC) method for integration will result in an unbiased estimator, it can have extremely high variance as most latent configurations do not explain a given observation well.
\citet{KingmaW13} and \citet{RezendeMW14} proposed a Variational Autoencoder (VAE) model that overcomes this challenge by using amortized variational inference.
VAEs maximize the evidence lower bound (ELBO) which is  derived by introducing an approximate posterior distribution \(q_\phi(z|x)\) (also known as \textit{recognition model}, \textit{inference network} or simply \textit{encoder}) parameterised as a neural network:
\begin{equation}
\label{eq:elbo}
\begin{split}
     \log{p_\theta(x)} \ge \mathbb{E}_{q_\phi(z|x)}\left[\log{p_\theta(x|z)}\right] & -  \\  \KLD{q_\phi(z|x)}{p(z)}&
\end{split} 
\end{equation}
The bound is maximized with respect to both generator parameters \(\theta\)
and encoder parameters \(\phi\).
Importantly,  gradients of the ELBO with respect to \textit{generator} parameters  \(\theta\) can be estimated with a relatively low variance.
Optimizing ELBO with respect to parameters \(\phi\) is equivalent to minimizing  the KL divergence between the true posterior and the encoder \(\KLD{q_\phi(z|x)}{p_\theta(z|x)}\).

\subsection{Posterior collapse}
Efficient and simple training of VAEs led to a surge of recent interest in 
deep generative modeling of structured data \citep{BowmanVVDJB16, Bombarelli15}.
However, prior work has established that VAE training often suffers from the {\it posterior collapse} problem, where the generative model learns to ignore a subset or all of the latent variables.
This phenomenon is more common when the generator \(p_\theta(x|z)\) is parametrised as a strong autoregressive model, for example, an LSTM \citep{HochreiterS97}.
A powerful autoregressive decoder has sufficient capacity to achieve good data likelihood without using latent code, thus intuitively having a KL term in the ELBO encourages the encoder to be independent of data \(q_\phi(z|x)\approx p(z)\).
While holding the KL term responsible for posterior collapse makes intuitive sense, the mathematical
mechanism of this phenomenon is not well understood, and spurious local optima that prefer the posterior-collapse solution may arise even during optimization of the exact marginal likelihood \citep{abs-1911-02469}.

\section{Our model}
In this section, we propose an alternative to a maximum likelihood estimation via kernel density estimation and discuss its properties.
Then we derive two upper bounds of KL divergence between the model and the nonparametric (KDE) distribution.
We discuss the pitfalls of the {na\"ive upper bound} and use an importance sampling distribution to derive a much tighter upper bound.
Finally, we describe how to use policy distillation approach to minimize the Levenstein reconstruction error.
If the reader is not interested in these derivations, they can skip to subsection \ref{subsec:surr}.

\subsection{Kernel density estimation}
Let us introduce a nonparametric probability distribution of sequences given a training set \(\mathcal{D}\), using a kernel density estimation (KDE) framework for discrete data \citep{aitchison1976multivariate}:
\begin{equation*} \label{nonp_pd}
    p_{\tau, \mathcal{D}}(x)=\frac{1}{|\mathcal{D}|}\sum_{x_k \in \mathcal{D}} \frac{\exp\left\{-\frac{1}{\tau}D_{\text{edit}}(x, x_k)\right\}}{Z_k}
\end{equation*}
Here \(D_{\text{edit}}\) stands for the Levenshtein distance \citep{levenshtein1966binary} between two sequences, \(Z_k\) denotes the normalizing constant of the kernel \(\sum_x\exp\left\{-\frac{1}{\tau}D_{\text{edit}}(x, x_k)\right\}\), and \(\tau\) is a smoothing parameter called the bandwidth of the kernel or, more appropriate in this case, temperature.
Intuitively, KDE estimates the probability of a particular sequence \(x\) by measuring how similar it is to the datapoints from the training dataset. 
If the tolerance for deviating from the original datapoint is infinitesimal, the empirical distribution is recovered: \(\lim_{\tau \to 0}p_{\tau, \mathcal{D}}(x)=\delta_{\mathcal{D}}(x)\).
Unfortunately, directly working with \(p_{\tau, \mathcal{D}}\) distribution is problematic. 
For example, evaluating the probability of a sequence \(x\) is infeasible due to the intractable partition functions.
It is straightforward to show that MLE optimization is equivalent to minimizing Kullback–Leibler divergence of model \(p_\theta(x)\) from empirical distribution \(\delta_\mathcal{D}(x)\).
Instead, we propose to optimize the reverse KL divergence:
\begin{align}
\nonumber
    \KLD{p_\theta(x)}{p_{\tau, \mathcal{D}}(x)} =& -\mathbb{E}_{p_\theta(x)}\left[\log p_{\tau,\mathcal{D}}(x)\right] \\ &- H(p_\theta(x))
    \label{eq:kde-loss}
\end{align}
Both directions of KL divergences are legitimate objective functions. 
The differences in their behaviour are well understood and used in approximate Bayesian inference \citep{Huszar15} and implicit generative models \citep{MohamedL16}.
The differences are the most striking when model underspecification is present, for example, trying to fit parameters of a small LSTM language model using massive dataset.
Minimizing forward \(\KLD{p_{\mathcal{D}}}{p_\theta}\) will lead to models that overgeneralise and produce samples that have low probability under \(p_{\mathcal{D}}\).
This is due to the mode covering behaviour of such loss, that requires all modes of \(p_{\mathcal{D}}\) to be matched by \(p_\theta\).
In other words, minimising forward KL will create a model that can produce all the behaviour that is observed in real data, at the cost of introducing behaviours that are never seen \citep{abs-1904-09751}.
On the other hand, using reverse \(\KLD{p_\theta}{p_{\mathcal{D}}}\) will yield a model that avoids any behaviour that is unlikely under \(p_{\mathcal{D}}\) at the cost of ignoring modes of the empirical distribution.
Usually, this reverse KL is optimised by employing the GAN framework.
The advantage of our proposed loss is that it is not as brittle and hard to train as GAN for text \citep{abs-1905-09922}, while enjoying the same theoretical benefits.

\subsection{Na\"ive upper bound}
The loss in Eq.~\ref{eq:kde-loss} cannot be optimized directly because of intractable \(p_{\tau, \mathcal{D}}\) distribution. 
One option is to apply the Jensen's inequality to the cross entropy term
to derive an upper bound:
\begin{gather}
\label{eq:naive_ub}
\begin{aligned}
    &\ \KLD{p_\theta(x)}{p_{\tau, \mathcal{D}}(x)} \leq - H(p_\theta(x)) + \text{const}\\ &+\frac{1}{\tau|\mathcal{D}|}\sum_{x_k\in\mathcal{D}}\mathbb{E}_{p_\theta(x)}\left[D_{\text{edit}}(x, x_k)\right] 
\end{aligned}
\raisetag{30pt}
\end{gather}
where \(\text{const}=\sum_{x_k \in \mathcal{D}}\log{Z_k}\).
Another interpretation of \(\tau\), namely that it controls the desired level of the entropy of the model, is apparent from the bound.
Despite the simplicity of this bound, it is too loose and cannot be used for learning.

\begin{figure}
    \centering
    \begin{subfigure}[b]{0.49\linewidth}
        \centering
        \includegraphics[width=\linewidth]{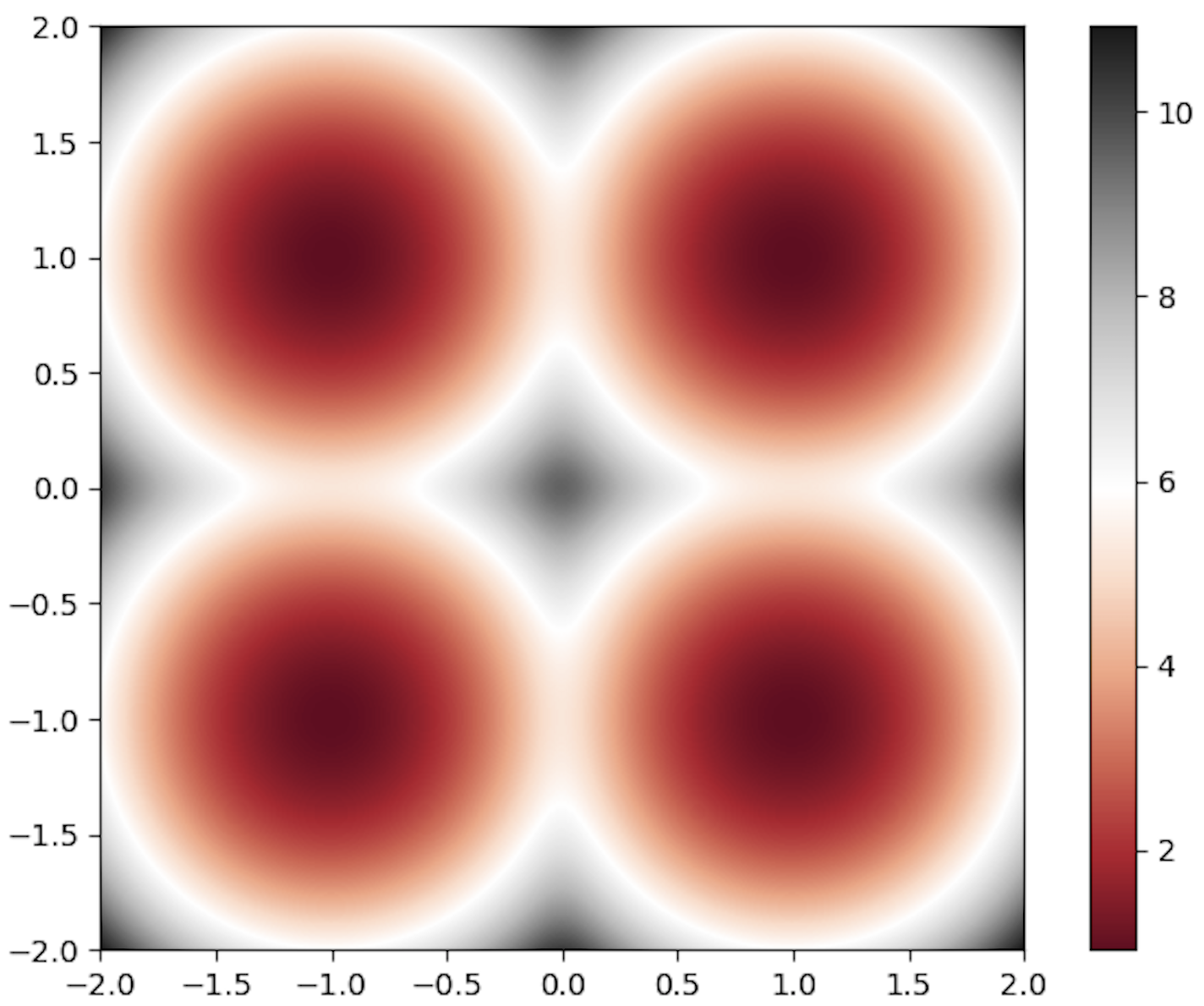}
        \caption{Original reward.}
        \label{fig:original_loss}
    \end{subfigure}
    \begin{subfigure}[b]{0.49\linewidth}
        \centering
        \includegraphics[width=\linewidth]{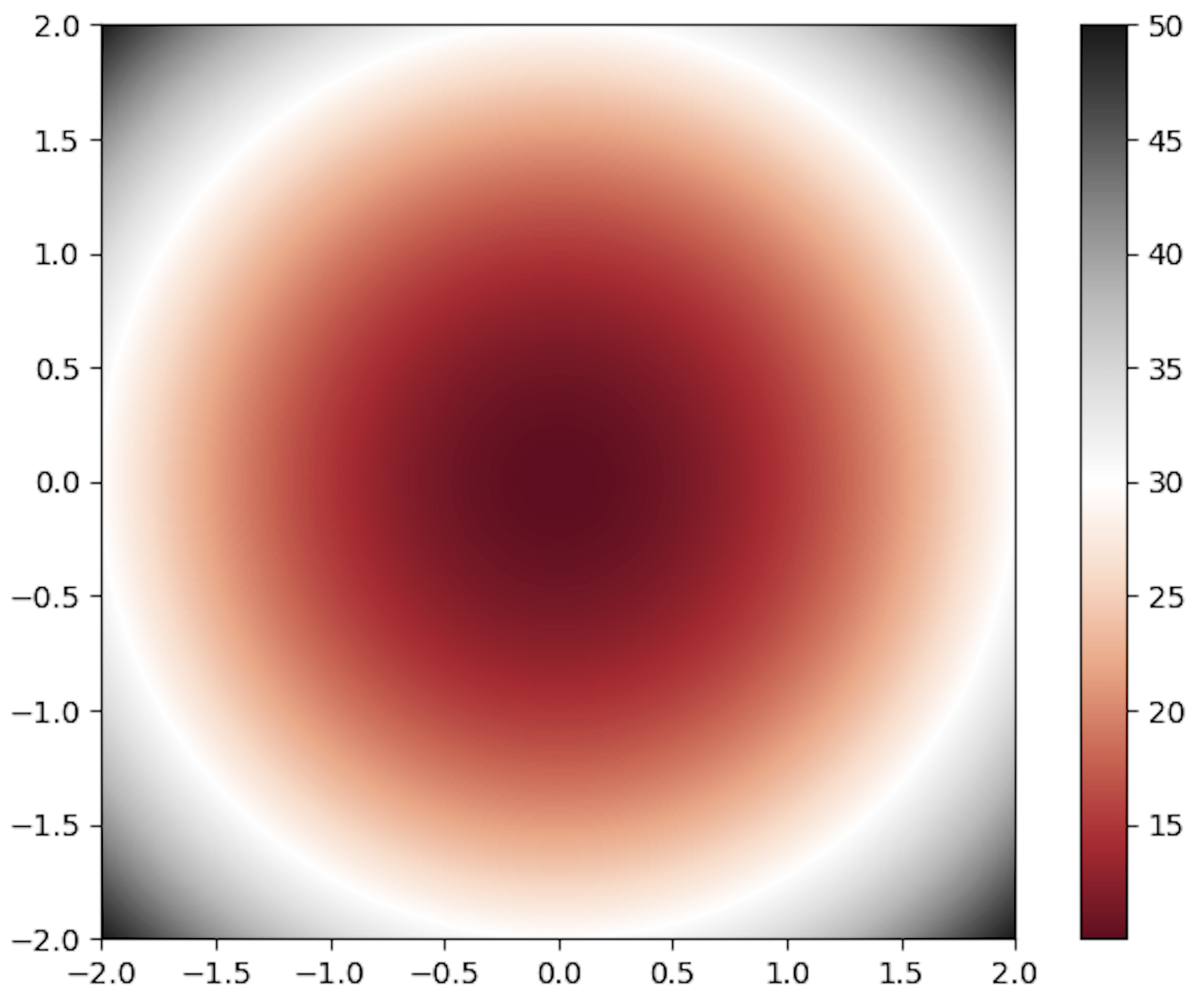}
        \caption{Na\"ive bound reward.}
        \label{fig:naive_bound}
    \end{subfigure}
    \caption{Reward functions associated with the original loss and na\"ive bound.}
\end{figure}

To obtain a better intuitive understanding of why it is the case, let's consider the continuous case of density estimation with the Euclidean squared distance as an ``edit'' distance.
In this setup, KDE distribution is simply equal to the mixture of Gaussians which we want to distil into our parametric model.
Let's assume a dataset contains four points than the contour lines of a corresponding reward function that the model \(p_\theta\) is trying to probabilistically optimize (excluding the entropy term) is shown in Figure \ref{fig:original_loss}.
The reward function associated with the na\"ive bound (Fig. \ref{fig:naive_bound}) assigns highest value to the part of the space that represents centre of mass of a dataset.
Obviously, this creates ``deceitful'' optimums and has catastrophic consequences for the optimized model. For discrete case this is especially evident if one uses the Hamming distance.
The Hamming distance is defined as the number of positions at which the corresponding symbols are different.
In other words, the Hamming distance between two sequences can be decomposed into a sum
of token distances: \(D_H(x, x_k)=\sum_i d(x^i, x_k^i)\), where $i$ is a position index and $d(x^i, x_k^i)$ is equal to $1$ if the tokens match and $0$, otherwise.
Thus, the summations over examples and over tokens in a sampled sequence can be swapped in the last term of the loose upper bound: \(\sum_{i}\mathbb{E}_{\theta}(x^i|x^{<i})\left[\sum_{x_k \in \mathcal{D}}d(x^i, x^i_k)\right]\).
This implies that optimizing this loss will result in matching the next token probability distribution with the empirical (position-specific) marginal distribution (centre of mass of a dataset), while disregarding the generated prefix.
With the Levenshtein distance, it is hard to determine precisely what would be the optimum of the bound, yet it is clear that it should not be used for learning.

\subsection{Tighter upper bound}
\paragraph{Importance sampling.}
Introducing an importance sampling (IS) distribution \(\gamma(k|x)\) is a viable way to make the bound tighter.
This distribution should tell us how important it is to minimise the distance between a generated sequence and the \(k\)-th example from the dataset.
Knowing that we want to build a latent variable model, we can substitute a sequence \(x\) with its corresponding latent code \(z\).
Further, we would like the model to be useful for representation learning, thus we will use an encoder \(q_\phi(z|x)\) to instantiate IS-distribution: \(\gamma(k|z)=\frac{1}{|\mathcal{D}|}\frac{q_\phi(z|x_k)}{q_\phi(z)}\), where \(q(z)=\frac{1}{|\mathcal{D}|}\sum_{x_i\in\mathcal{D}}q_\phi(z|x_i)\), which in variational inference literature is known as the aggregated posterior\footnote{It is worth mentioning that even though the introduced encoder is reminiscent of an approximate posterior distribution, the model is not trying to explicitly minimize the difference between the encoder and the true posterior.} \citep{MakhzaniSJG15}.
Intuitively, \(\gamma(k|x)\) measures compatibility between $z$ (i.e. the latent representations of $x$) and the data point $x_k$.
Applying a Jensen's inequality to  the loss (Eq.~\ref{eq:kde-loss}) while using  IS-distribution \(\gamma(k|z)\), results in the following upper bound (see details in Appendix~\ref{appendix_ub}):
\begin{gather}
\label{eq:true_bound}
\begin{aligned}
    &\KLD{p_\theta(x)}{p_{\tau, \mathcal{D}}(x)} \le \mathcal{L_{\phi,\theta}(\mathcal{D})} = \\
    & \frac{1}{\tau|\mathcal{D}|}\sum_{x_k \in \mathcal{D}}\mathbb{E}_{q_\phi(z|x_k)}\left[\frac{p(z)}{q_\phi(z)}\mathbb{E}_{p_\theta(x|z)}\left[D_{\text{edit}}(x, x_k)\right]\right]\\ 
    &+\frac{1}{|\mathcal{D}|}\sum_{x_k \in \mathcal{D}}\mathbb{E}_{q_\phi(z|x_k)}\left[\frac{p(z)}{q_\phi(z)}\log q_\phi(z|x_k)\right]\\ 
    &-\mathbb{E}_{p(z)}\log q_\phi(z)\\
    &+\frac{1}{|\mathcal{D}|}\sum_{x_k \in \mathcal{D}}\mathbb{E}_{q_\phi(z|x_k)}\left[\frac{p(z)}{q_\phi(z)}\log{Z_k}\right] - H(p_\theta(x))
\end{aligned}
\raisetag{80pt}
\end{gather}
This objective has a crucial property of scalability, as it is decomposed into a sum of instance-level losses that can be evaluated independently of each other. 
Hence, ubiquitous stochastic optimization can be used to scale the learning to massive datasets \citep{bottou2010large}.

\paragraph{Approximate bound.}
If the ratio \(\frac{p(z)}{q_\phi(z)}\) had been equal to one all parts of the bound would have gained clear meanings.
Namely, the first term becomes expected reconstruction error measured by the edit distance and the second term turns into the negative entropy of the encoder averaged across all inputs.
For this reason, we decided to consider constrained optimization task with \(q_\phi(z)=p(z)\) as a constrain. Using a method of Lagrange multipliers we can simply add \(\lambda\KLD{q_\phi(z)}{p(z)}\) term to the Eq. \ref{eq:true_bound}.
Having such regularization expression, we can approximate the terms in the bound by considering \(\frac{p(z)}{q_\phi(z)}=1\).
Assuming that Lagrange multiplier \(\lambda=1\) the next loss function is derived:
\begin{gather}
\label{eq:LVAE-loss}
\begin{aligned}
    &\mathcal{L_{\phi,\theta}(\mathcal{D})} \approx \\
    & \frac{1}{\tau|\mathcal{D}|}\sum_{x_k \in \mathcal{D}}\mathbb{E}_{q_\phi(z|x_k)}\left[\mathbb{E}_{p_\theta(x|z)}\left[D_{\text{edit}}(x, x_k)\right]\right]\\ 
    &+\frac{1}{|\mathcal{D}|}\sum_{x_k \in \mathcal{D}}\KLD{q_\phi(z|x_k)}{p(z)}\\ 
    &\frac{1}{|\mathcal{D}|}\sum_{x_k \in \mathcal{D}}\log{Z_k} - H(p_\theta(x))
\end{aligned}
\raisetag{80pt}
\end{gather}
The gradient of this approximate constrained bound will result in the biased gradients of the original bound. However, if the constraint is satisfied the bias disappears.

\subsection{Optimal Transport interpretation.}
\begin{figure}[h]
\centering
  \includegraphics[width=0.7\linewidth]{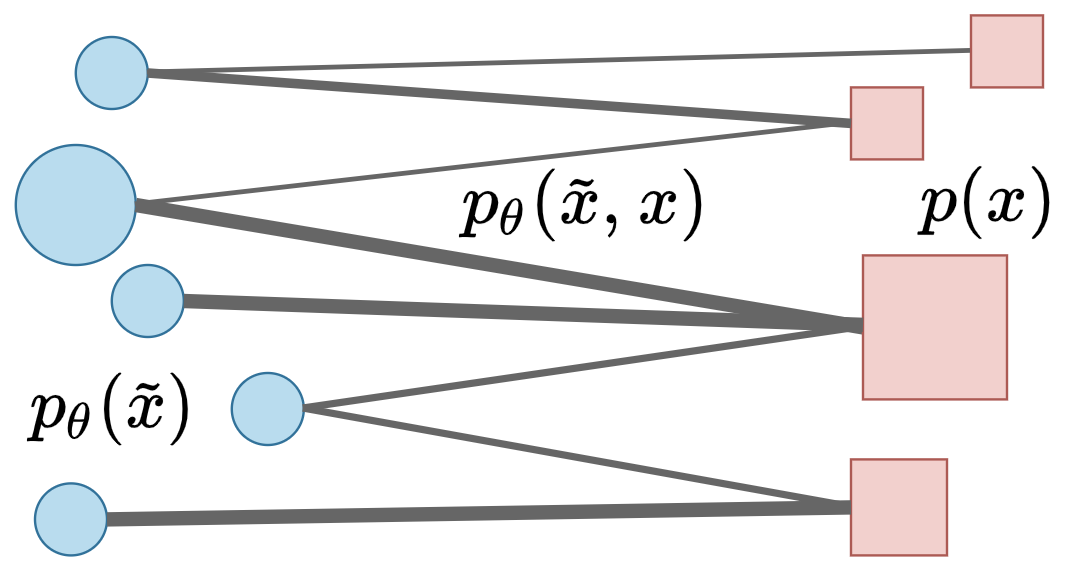}
  \caption{Empirical and model distributions are represented by squares and circles respectively. The area of the markers is proportional to the probability of each datapoint. Distances between markers correspond to the edit distance between datapoints. The thickness of lines is proportional to the corresponding value of the joint probability.}
  \label{fig:ot}
\end{figure}

Interestingly enough, it is possible to obtain a loss identical to Eq. \ref{eq:LVAE-loss} by considering an optimal transport task.
Let's inspect a Monge–Kantorovich transportation problem which in our case can be described as a task of pushing model probability mass \(p_\theta(\tilde{x})\) into an empirical distribution \(p(x)\):
\begin{equation}
\begin{split}
&\min_{\theta}\left\{\mathbb{E}_{p_\theta(x, \tilde{x})}\left[c(x, \tilde{x})\right]\right\} \\
&s.t. \sum_x p_\theta(x,\tilde{x})=p_\theta(\tilde{x}), \sum_{\tilde{x}} p_\theta(x, \tilde{x})=p(x).
\end{split}
\label{eq:ot}
\raisetag{40pt}
\end{equation}
The cost of transporting sampled datapoint \(\tilde{x}\) to the true datapoint \(x\) is given and in this case defined by the Levenshtein distance \(c(\tilde{x}, x )=D_{\text{edit}}(\tilde{x}, x)\).
The transportation plan can be represented by the joint distribution \(p_\theta(x, \tilde{x})\) (see Fig. \ref{fig:ot}) which indicates how much mass of \(\tilde{x}\) should be transported to the point \(x\).
This plan has to satisfy the mass conservation constraints mentioned in Eq. \ref{eq:ot}.
Following the same reasoning as in \citet{TolstikhinBGS18}, we will connect \(x\) and \(\tilde{x}\) through the latent space: \(p_\theta(x, \tilde{x}) = p_\theta(\tilde{x}|x)p(x)=\int p_\theta(\tilde{x}|z)q_\phi(z|x)p(x)dz\).
The mass conservation constraint for the \(p(x)\) is trivially satisfied by construction.
Knowing that a generative model \(p_\theta(x)=\int p_\theta(x|z)p(z)dz\) is a latent variable model, mass conservation constraint is equivalent to the constraint of prior distribution being equal to the aggregated posterior \(p(z)=q(z)\overset{def}{=}\sum_x q_\phi(z|x)p(x)\). The optimal transport loss can be derived using a method of Lagrange multipliers:
\begin{equation}
\mathbb{E}_{p(x)}\mathbb{E}_{q_\phi(z|x)}\mathbb{E}_{p_\theta(\tilde{x}|z)}\left[c(x, \tilde{x})\right] + \lambda \mathcal{D}(p(z), q(z))
\end{equation}
Where \(\mathcal{D}(p(z), q(z))\) is any divergence between two probability distributions.
The approximate upper bound Eq. \ref{eq:LVAE-loss} can be produced by selecting corresponding KL-divergence and adding the entropy regularization.
In other words, our approximate bound minimizes the transportation cost between a latent variable model and an empirical distribution while maintaining a high entropy for the former distribution.
This interpretation once again resonates with GAN models which can also be trained using a dual formulation of the Monge-Kantorovich task.

\subsection{Reconstruction error minimization.}
Inspecting the approximation of the KL upper bound (Eq.~\ref{eq:LVAE-loss}) one can notice that the last three elements can be effortlessly optimised.
For example, for some distributions (e.g. normal with diagonal covariance matrix) gradients of the KL can be computed analytically, or MC estimated with relatively low variance.
Even though the entropy of the generative model itself is intractable to compute, it is possible to derive a tractable upper bound (using Jensen's inequality).
Its gradients can be used for the entropy maximization.
The third term is a constant with respect to parameters, so can be safely ignored altogether.
On the other hand, the first term is the most challenging. 
This term tries to embed an input sentence \(x_k\) into a latent space \(z\) in a way that makes it easy to reconstruct the sentence back by keeping the edit distance to the original sequence as low as possible.
The inner part \(\mathbb{E}_{p_\theta(x|z)}\left[D_{\text{edit}}(x, x_k)\right]\) is especially difficult to optimise.

\paragraph{RL approach.}
This reconstruction loss can be seen as a reinforcement learning task and optimised with different variations of the REINFORCE algorithm \citep{Williams92}.
Tokens generated so far correspond to the agent's state, the agent's action corresponds to the symbol that should be produced next.
Despite the recent successes of deep RL, obtaining acceptable levels of performance  often requires an almost prohibitively large amount of experience to be acquired by the agent \citep{SalimansHCS17}.
Further, in the VAE setup, an encoder is usually optimised using low variance pathwise derivatives \citep{RezendeMW14}.
Even though an encoder in our method can be reparametrised, the pathwise derivative cannot be derived because there is no analytical expression for the inner term.

\paragraph{Policy distillation approach.}
If the expert policy exists, the learning efficiency of the RL task can be drastically improved.
\citet{SabourCN19} noticed that for the Levenshtein distance 
dynamic programming can be used to efficiently compute such expert policy, which they call the optimal completion (OC) policy \(\pi_{\text{OC}}(\cdot|\tilde{x}_{<t}, x)\).
In other words, given the generated prefix \(\tilde{x}_{<t}\) at time step \(t\) and the original sequence \(x\), we can determine what symbol should be generated to minimise the edit distance.
In NLP, such expert policies are known as dynamic oracles \citep{GoldbergN12}.
Figure~\ref{tab:ocp} contains an example of targets provided by such oracle for a particular generated sequence given a target sentence.
\begin{figure}[t]
\centering
\setlength\tabcolsep{4pt}
\scalebox{0.7}{
\begin{tabular}{lccccccc}
\toprule
Target & \(\texttt{<S>}\) & \(\texttt{this}\) & \(\texttt{pizza}\)  & \(\texttt{is}\) & \(\texttt{very}\) &  \(\texttt{good}\) & \(\texttt{<}\backslash \texttt{S>}\) \\ 
Generated & \(\texttt{<S>}\)& \(\texttt{the}\) & \(\texttt{risotto}\)  & \(\texttt{is}\) & \(\texttt{very}\) &  \(\texttt{good}\) & \(\texttt{<}\backslash \texttt{S>}\)\\ 
OC targets & \(\texttt{this}\) & \begin{tabular}{@{}c@{}}\(\texttt{\{this,}\) \\ \(\texttt{ pizza\}}\)\end{tabular}  & \begin{tabular}{@{}c@{}}\(\texttt{\{this,}\) \\ \(\texttt{pizza,}\) \\ \(\texttt{is}\}\)\end{tabular}  & \(\texttt{very}\) & \(\texttt{good}\) &  \(\texttt{<}\backslash \texttt{S>}\) & \\
\bottomrule
\end{tabular}
}
\caption{The optimal next words for each prefix based on the Levenshtein distance.} 
\label{tab:ocp}
\end{figure}
There is great interest in the RL field in methods that enable knowledge transfer to agents based on already trained policies \citep{RusuCGDKPMKH15} or human examples \citep{PieterN04}.
One of the most successful techniques for knowledge transfer is policy distillation \citep{RossGB11}, where an agent is trained to match the state-dependent probability distribution over actions provided by a teacher/oracle. We use policy distillation
as a proxy to optimizing the term \(\mathbb{E}_{p_\theta(x|z)}\left[D_{\text{edit}}(x, x_k)\right]\) in Eq.~\ref{eq:LVAE-loss}.

Usually, policy distillation is done by following updates in the gradient-like direction:
\begin{equation}
\label{eq:distillation}
    \mathbb{E}_{\tilde{x} \sim \rho} \left[
    \sum_{t=1}^{|\tilde{x}|}\! \nabla_{\theta}\KLD{\pi_{\text{oc}}(\cdot|\tilde{x}_{<t}, x)}{p_{\theta}(\cdot|\tilde{x}_{<t}, z)} 
    \right]
\end{equation}
The distribution \(\rho\) is known as a control policy \citep{CzarneckiPOJSJ19}, and it is used to generate trajectories over which the distillation process is performed.

\paragraph{OC control policy.}
A variety of different control policies can be used.
For example, if the control policy is equal to the teacher policy, Eq.~\ref{eq:distillation} leads to a method which is known as teacher forcing.
Interestingly enough, using the OC policy as the control policy in Eq.~\ref{eq:LVAE-loss}, recovers the ELBO objective (Eq.~\ref{eq:elbo}).

\paragraph{Generator as a control policy.}
Generally, better results are obtained by training a model under its own predictions\footnote{It is worth mentioning that in this case, distillation updates do not form a gradient field anymore.
Nonetheless, it does not pose a real problem (for details see \citet{CzarneckiPOJSJ19}).} \(\rho=p_\theta\).
In our experiments, we observed even better results with a distribution that assigns all the probability mass to the greedy argmax of the model \(\delta_{p_\theta}(x)=\mathbbm{1}_{\gargmax p_\theta}(x)\).
\paragraph{A mixture control policy}
Important advantage of adopting the policy distillation approach is that, as latent code \(z\) can be expressed as a deterministic function \(g_\phi(x, \epsilon)\) of parameters, input and independent noise, nothing prevents us from employing the reparameterization trick~\cite{KingmaW13} for the encoder.
Additionally, the distillation process implicitly includes a model entropy term from Eq.~\ref{eq:LVAE-loss}.
For example, in case of a multimodal teacher distribution, the KL term in Eq.~\ref{eq:distillation} will have extreme values if a model picks only one mode.
Hence the distillation process keeps the model away from the mode-collapsing solutions.
Using the latent variable is the only way our model can access a gold standard prefix.
For that reason, if the latent space is small, the optimisation problem becomes extremely hard.
Another consequence of a small latent space is that the IS distribution will not be able to properly partition the dataset, thus the upper bound will become too loose to be useful for learning.
To overcome this issue we use a mixture of the previously introduced greedy argmax and teacher distributions as the control policy in Eq.~\ref{eq:distillation}.

\paragraph{Surrogate loss and algorithm.}
\label{subsec:surr}
Despite the seemingly complex derivation of the method, proposed technique is quite simple and easy to use in practice.
To use the proposed method within any automatic differentiation framework, it is sufficient to construct the following surrogate loss:
\begin{gather}
\label{eq:final_gradients}
\begin{aligned}
    &\lambda\hat{\mathbb{E}}_{\tilde{x} \sim \delta_{p_\theta}} \left[\sum_{t=1}^{|\tilde{x}|} \KLD{\pi_{\text{OC}}}{p_{\theta}(\cdot|\tilde{x}_{<t}, g_\phi(x_k, \epsilon))} \right]\\
    &-\alpha\sum_{t=1}^{|x_k|}\log p_\theta(x^k|x^k_{<t}, g_\phi(x_k, \epsilon)))\\
    &+\tau\KLD{q_\phi(z|x_k)}{p(z)}
\raisetag{42pt}
\end{aligned}
\end{gather}
A surrogate objective is a function of the inputs that, once differentiated, gives an unbiased gradient estimator of the original loss \citep{SchulmanHWA15}.
Here \(\hat{\mathbb{E}}\) denotes the empirical average, which means that one does not need to backpropagate through the sampling distribution.
Hyperparameters \(\lambda\) and \(\alpha\) determine, correspondingly, the level of mixture of student and teacher policies  in the control policy.
Algorithm \ref{alg:LEV_VAE} outlines the optimization procedure for our method.
\begin{algorithm}[t]
\SetKwInOut{Input}{Input}
\SetKwInOut{Output}{Output}
 \Input{Model \(x, z \sim p_\theta(\cdot|z), p(z)\) \newline
IS distribution \(q_\phi(z|x)\)\newline
dataset \(\mathcal{D}\)}
 \Output{parameters \(\theta, \phi\)}
 Initialize \(\theta, \phi\) randomly.

 \While{not converged}{
  \vspace{0.25ex}
  Sample datapoint \(x_k \sim \mathcal{D}\) \\
  Sample latent code \(z_k \sim q_\phi(z|x_k)\) \\
  Generate \(\tilde{x}\) from the model \(p_\theta(\cdot|z_k)\)\\ 
  Store corresponding \(p_\theta(\cdot|\tilde{x}_{<t}, z_k)\) \\
  Compute OC policy for Levenshtein distance using dynamic programming: \(\prod_{t=1}^{|\tilde{x}|}\pi_{\text{OC}}(\cdot|\tilde{x}_{<t},x_k)\) \\
  Compute \(\nabla_{\theta,\phi} \KLD{\pi_{\text{OC}}}{p_\theta(\cdot|\tilde{x}_{<t}, z_k)}\) \\
  Compute \(\nabla_{\theta,\phi} -\log p_\theta(x_k|z)\) \\
  Compute \(\nabla_{\phi} \KLD{q_\phi(z|x_k)}{p(z)}\) \\
   Update \(\theta, \phi\) using SGD. \\
 }
 \caption{Levenshtein VAE}
 \label{alg:LEV_VAE}
\end{algorithm}

It is worth emphasising once again, that Eq.~\ref{eq:LVAE-loss} cannot be minimised unless some information is stored in the latent code.
The generator does not have any access to the true sequence except through the latent code, and because the objective is to minimise the edit distance between the input and generated sequences, it has no other choice but to store information in the latent space.

\section{Related work}
\label{sec:related_work}
Since the posterior collapse issue was observed, several methods have been proposed to mitigate it.
They can be roughly divided into four groups.
The first contains models that modify the architecture of the model, 
for example, by weakening a decoder \citep{BowmanVVDJB16, YangHSB17} or by introducing additional connections that enforce strong links between the latent variables and the likelihood function \citep{DiengKRB19}.
The second is represented by methods that introduce more flexible approximate posterior or prior distributions \citep{TomczakW18, abs-1904-08194} or fix some of their parameters and thus putting a cap on a minimal KL value \citep{XuD18}.
The third group consists of methods that modify the amortized inference network optimization procedure of VAE \citep{HeSNB19, KimWMSR18}. 
Finally, the fourth group contains models that modify ELBO objective by introducing an additional loss that encourages high mutual information between observed and latent variables \citep{ZhaoZE17, ZhaoSE17b, HigginsMPBGBML17, abs-1904-08194}.
Our model is most closely related to the last group.
Even though proposed approach is not directly connected to the ELBO, it still can be reinterpreted as a model that augments ELBO with the term that encourages high mutual information.

\section{Experiments}
\begin{table*}[t!]
\centering
\setlength\tabcolsep{4pt}
\small
\begin{tabular}{lcccccc}
\toprule
\textbf{Method}& \textbf{Lev. D} & \textbf{-ELBO} & \textbf{Recon.} & \textbf{KL} & \textbf{PPL}  & \textbf{-LL} \\ \midrule
LSTM-LM                          & --     & --     & --     & --    & 72.22  & 104.67 \\
VAE                              & 0.90   & 105.00 & 104.99 & 0.01  & 73.10  & 104.99 \\
\(\beta\)-VAE\(_{\beta=0.6}\)                    & 0.85   & 105.09 & 101.73 & 3.36  & 72.38  & 104.74 \\
Cyc. ann. VAE\(_{M=5}\)                    & 0.86   & 104.80 & 103.40 & 1.40  & 72.07  & 104.64 \\
Lev. VAE\(_{\alpha=1.0,\lambda=0.15}^{\tau=1.0}\) & 0.87   & 104.72 & 103.42 & 1.30  & 72.11  & 104.65 \\
Lev. VAE\(_{\alpha=1.0.\lambda=1.0}^{\tau=1.0}\) & 0.79   & 109.74 & 98.01 & 11.73 & 80.19  & 107.25 \\
\bottomrule
\end{tabular}
\caption{Results on PTB test set for various methods.} 
\label{tab:ptb}
\end{table*}

The effectiveness of our method is validated by obtaining comparable results to other methods for the language modelling task and achieving better results in representation learning as measured by probing classifiers.
We conducted experiments on three different datasets: Penn Treebank dataset \citep{MarcusSM94} for language modeling, SNLI \cite{BowmanAPM15} and Yelp \citep{ShenLBJ17} datasets  for representation learning evaluation.
The  code to reproduce the experimental results is publicly available.\footnote{\url{https://github.com/anonymised_link}}

\subsection{Penn Treebank}
We compared our method with several baselines, including the standard RNN language model and VAE-based language models: the vanilla VAE model, \(\beta\)-VAE \citep{HigginsMPBGBML17} and cyclical annealing VAE \citep{FuLLGCC19}.
We started by  determining the optimal hyperparameters for the LSTM language model, which we used within generators of all models.
A random search \citep{BergstraB12} was used for hyperparameters optimisation. Appendix \ref{sec:hyp_app} contains all the details of this process and the discovered optimal values. 
An encoder for the VAE-based methods was implemented as a one-layer bidirectional LSTM.
A Gaussian approximate posterior with \(32\)-dimensional diagonal covariance matrix was used for all models.
The generator was implemented as a vanilla LSTM language model with the first hidden state initialised by the projection from the latent space.
To achieve a fair comparison of methods, we tuned additional hyperparameters for each algorithm.
Particularly, annealing cycle period \(M\) measured in epochs for cyclical annealing VAE and KL weight \(\beta\) for \(\beta\)-VAE.
For our method we fixed KL weight \(\tau=1\) and teacher forcing weight \(\alpha=1\) and tuned only optimal completion policy weight~\(\lambda\).

The results are in Table \ref{tab:ptb}.
\(\text{Lev. D}\) denotes the average Levenshtein distance between a sequence from the dataset and its  reconstruction. 
Our method performs competitively on this dataset.
For all VAE models negative log-likelihood (\(\text{-LL}\)) and perplexity (PPL) was estimated by importance sampling using the trained approximate posterior \(q(z|x)\) as the importance distribution with 1000 samples.

\subsection{Probing the latent space}
\subsubsection{Yelp sentiment analysis}
To quantify representation learning capability of the proposed method, we trained our model on Yelp sentiment dataset preprocessed by \citet{ShenLBJ17}.
Then we trained a linear classifier to predict sentiment labels relying on the representation obtained by the approximated posterior distribution, namely expected value \(\mu\).
\begin{table}[t!]
\centering
\small
\setlength\tabcolsep{4pt}
\begin{tabular}{lcccccc}
\toprule
\textbf{Method}& \textbf{Lev.D} & \textbf{Recon.} & \textbf{KL}    & \textbf{Acc.\%} \\ \midrule
AE                            & 0.00  & 0.03 & -- &  86.57  \\
L. AE                         & 0.00  & 0.08 & -- &   89.93  \\
\(\beta\)-VAE\(_{\beta=1.0}\) / VAE           & 0.79 & 31.84 & 0.01    & 91.90 \\
Cyc. ann. VAE\(_{M=18}\)             & 0.80 & 32.48 & 0.02   & 91.95 \\
L. VAE\(_{\alpha=1.0,\lambda=1.0}^{\tau=0.7}\) & 0.30 & 10.06 & 33.15   & 92.66 \\
L. VAE\(_{\alpha=0.0,\lambda=1.0}^{\tau=0.2}\)    & 0.18  & 7.44 & 49.95   & 92.91  \\
\bottomrule
\end{tabular}
\caption{Results on Yelp test set for various methods.}
\label{tab:yelp}
\end{table}

We fixed latent space dimensionality at \(256\).
Again, to make a fair comparison, we tuned hyperparameters of each method (for details, see appendix \ref{sec:hyp_app}).
The model selection process was based on the linear classifier performance on the validation set.
For our method we investigated two distinct configurations: with (\(\alpha=1, \lambda=1\)) and without (\(\alpha=0, \lambda=1\)) teacher forcing and tuned KL weight \(\tau\).
Also, a simple non-variational Levenshtein autoencoder (\(\alpha=0,\lambda=1, \tau=0\)) was trained.
Probably due to the nature of the sentiment analysis task, the good performance among baseline VAE methods was achieved by the models with tiny KL values.
However, the variations of our model achieved better results and have much bigger KL between the prior and the approximate posterior.
One can see from Table~\ref{tab:yelp}, \(\text{Lev. AE}\) row, that our method does not merely perform latent space regularization as bag-of-words-augmented VAEs \citep{ZhaoZE17}: for it can almost perfectly autoencode each sentence from the test set.

\paragraph{Latent space}
To qualitatively show the difference between our method and \(\beta\)-VAE, we selected two models with similar KL values (\(19.35\) for our model and \(19.29\) for \(\beta\)-VAE).
For each sentence in the test set, we generated its reconstruction using a latent code and  argmax decoding.
Then for each position in the original sentence, we checked whether it is present in the reconstruction.

\begin{figure}[t!]
  \includegraphics[width=\linewidth]{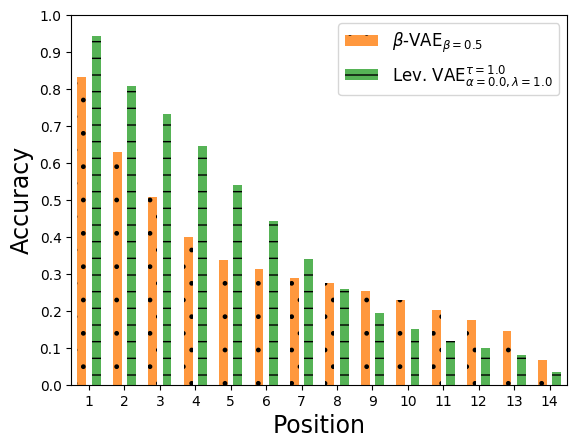}
  \caption{Word reconstruction accuracy depending on the position in the input sentence.}
  \label{fig:rec}
\end{figure}
As one can see from Figure \ref{fig:rec}, \(\beta\)-VAE accuracy decreases with a higher rate in comparison to our model.
We argue that this is because our model does not have direct access to a gold standard prefix and can only use the information available in the latent code, hence it must learn how to utilize it to generate correct reconstructions.
In contrast, \(\beta\)-VAE overrelies on the prefix and cannot at test time recover from its own mistakes. 

\subsubsection{Natural language inference}
Observing very low KL values for the best baseline methods on Yelp dataset, we decided to evaluate our model 
on a harder task,  natural language inference (NLI) task using SNLI dataset \citep{BowmanAPM15}.
NLI consists of predicting the relationship between two sentences which can be either entailment, contradiction, or neutral.
The task can be formulated as a three-way classification problem.
First of all, we pretrained our model using language modelling task on SNLI dataset with deduplicated premises.
Then we trained a linear probe using a feature vector obtained by concatenating vectors \(\mu_{pre}, \mu_{hyp}, \mu_{pre} - \mu_{hyp}\) and \(\mu_{pre} \odot \mu_{hyp}\), where \(\mu\) is expected value of the approximate posterior distribution.
\begin{table}[t!]

\centering
\small
\setlength\tabcolsep{4pt}

\begin{tabular}{lccccc}
\toprule
\textbf{Method}& \textbf{Lev.D} & \textbf{Recon.} & \textbf{KL} &  \textbf{Acc.\%} \\ \midrule
AE                                             & 0.02 & 1.23 & -- &  59.71  \\
L. AE                                          & 0.02 & 1.49 & -- &  61.49  \\
VAE                                            & 0.70 & 31.13 & 0.01  & 43.36 \\
\(\beta\)-VAE\(_{\beta=0.1}\)                  & 0.09 & 2.49 & 54.91  & 60.33 \\
Cyc. ann. VAE\(_{M=2}\)                       & 0.69 & 30.60 & 0.57  &  55.54 \\
L. VAE\(_{\alpha=0.0,\lambda=1.0}^{\tau=0.4}\) & 0.21 & 13.44 & 31.93 & 63.31  \\
L. VAE\(_{\alpha=1.0,\lambda=1.0}^{\tau=0.6}\) & 0.17 & 7.37 & 36.14  & 63.80 \\
\bottomrule
\end{tabular}
\caption{Results for SNLI test set for various baselines.}
\label{tab:snli}
\end{table}



From Table \ref{tab:snli}, one can see that even the simple non-variational Levenshtein autoencoder performs better than the baselines.
If we add more structure to the latent space with the KL regularization, our method achieves even better results.
\begin{figure}[h]
  \includegraphics[width=\linewidth]{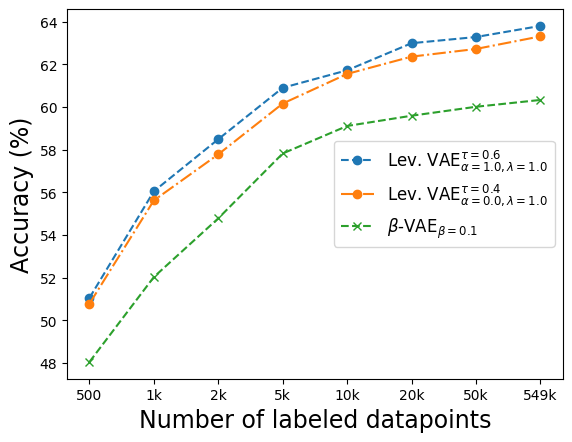}
  \caption{Accuracy of the linear classifier with a different size of a training dataset.}
  \label{fig:semi-s-accuracy}
\end{figure}
We also emulated a semi-supervised setup by varying the amount of labeled data for the classifier  (Figure \ref{fig:semi-s-accuracy}).
Our method outperforms \(\beta\)-VAE in all regimes.

\section{Discussion}
In this work, we focus on the Levenshtein distance, yet other distances can be used, assuming that their corresponding OC policies are easy to compute.
Scheduled sampling \citep{BengioVJS15}, a popular technique for avoiding exposure bias problem~\cite{RanzatoCAZ15}, can be reinterpreted as an optimal completion policy distillation algorithm for the Hamming distance.
In fact, they minimise the na\"ive upper bound (Eq.\ref{eq:naive_ub}) of the proposed loss, as a consequence the method has the same issues that come with this loose bound.

Besides posterior collapse, another serious issue affecting VAEs, as well as
any other models trained using teacher forcing, is \textit{exposure bias}~\cite{RanzatoCAZ15}. The models that have never been trained on their own errors may not be robust to them at test time. In our approach, the model generates sequences at train time and there is no such mismatch between test and train regimes. We leave investigation of the degree to which it can improve generation quality to future work.



\section{Conclusion}
In this paper, we have introduced a novel method for learning latent variable models that minimises the KL divergence of the KDE nonparametric distribution from the model.
It is simple to use and efficient and does not suffer from the posterior collapse issue.
We demonstrated its effectiveness for representation learning and language modelling.


\bibliography{acl2020}
\bibliographystyle{acl_natbib}

\clearpage
\appendix
\section{Upper bound}
\label{appendix_ub}
\begin{equation*}
\begin{split}
    \KLD{p_\theta(x)}{p_{\tau, \mathcal{D}}(x)} =& -\mathbb{E}_{p_\theta(x)}\left[\log p_{\tau,\mathcal{D}}(x)\right] \\ &- H(p_\theta(x))
\end{split} 
\end{equation*}
Let's consider the cross entropy term:
\begin{equation*}
\resizebox{.5\textwidth}{!}{%
  $
\begin{split}
    &\mathbb{E}_{p_\theta(x)}\left[\log p_{\tau,\mathcal{D}}(x)\right] = \mathbb{E}_{p(z)}\mathbb{E}_{p_\theta(x|z)}\left[\log p_{\tau,\mathcal{D}}(x)\right] = \\ & \mathbb{E}_{p(z)}\mathbb{E}_{p_\theta(x|z)}\left[\log\left[\frac{1}{|\mathcal{D}|}\sum_{x_k \in \mathcal{D}}\frac{\gamma(k|z)}{\gamma(k|z)} \frac{\exp\left\{-\frac{1}{\tau}D_{\text{edit}}(x, x_k)\right\}}{Z_k}\right]\right]
\end{split} $}
\end{equation*}
Applying Jensen's inequality one can derive: 
\begin{equation*}
\begin{split}
    &-\mathbb{E}_{p_\theta(x)}\left[\log p_{\tau,\mathcal{D}}(x)\right] \le \\ & \frac{1}{\tau}\mathbb{E}_{p(z)}\mathbb{E}_{p_\theta(x|z)}\left[\sum_{x_k \in \mathcal{D}}\gamma(k|z) D_{\text{edit}}(x, x_k)\right] + \\ &\mathbb{E}_{p(z)}\mathbb{E}_{p_\theta(x|z)}\left[\sum_{x_k \in \mathcal{D}}\gamma(k|z)\log\gamma(k|z)\right] + \\ &\mathbb{E}_{p(z)}\mathbb{E}_{p_\theta(x|z)}\left[\sum_{x_k \in \mathcal{D}}\gamma(k|z)\left(
    \log|\mathcal{D}| + \log{Z_k}\right)\right]
\end{split} 
\end{equation*}
After expanding \(\gamma(k|z)\) it is evident that all elements in the second and third terms do not depend on \(x\). After exchanging order of integration and summation:
\begin{equation*}
\resizebox{.5\textwidth}{!}{%
  $
\begin{split}
    &-\mathbb{E}_{p_\theta(x)}\left[\log p_{\tau,\mathcal{D}}(x)\right] \le \\ 
    & \frac{1}{\tau|\mathcal{D}|}\sum_{x_k \in \mathcal{D}}\mathbb{E}_{p(z)}\left[\frac{q_\phi(z|x_k)}{q_\phi(z)}\mathbb{E}_{p_\theta(x|z)}\left[D_{\text{edit}}(x, x_k)\right]\right] + \\ 
    &\frac{1}{|\mathcal{D}|}\sum_{x_k \in \mathcal{D}}\mathbb{E}_{p(z)}\left[\frac{q_\phi(z|x_k)}{q_\phi(z)}\left(\log q_\phi(z|x_k) - \log q_\phi(z) +\log{Z_k}\right)\right]
\end{split} $}
\end{equation*}
Simplifying further, the next upper bound for the cross entropy is obtained:
\begin{equation*}
\begin{split}
    &-\mathbb{E}_{p_\theta(x)}\left[\log p_{\tau,\mathcal{D}}(x)\right] \le \\
    & \frac{1}{\tau|\mathcal{D}|}\sum_{x_k \in \mathcal{D}}\mathbb{E}_{q_\phi(z|x_k)}\left[\frac{p(z)}{q_\phi(z)}\mathbb{E}_{p_\theta(x|z)}\left[D_{\text{edit}}(x, x_k)\right]\right] + \\ 
    &\frac{1}{|\mathcal{D}|}\sum_{x_k \in \mathcal{D}}\mathbb{E}_{q_\phi(z|x_k)}\left[\frac{p(z)}{q_\phi(z)}\log q_\phi(z|x_k)\right] - \\ 
    &\mathbb{E}_{p(z)}\left[\frac{\frac{1}{|\mathcal{D}|}\sum_{x_k \in \mathcal{D}}q_\phi(z|x_k)}{q_\phi(z)}
    \log q_\phi(z)\right] +\\
    &\frac{1}{|\mathcal{D}|}\sum_{x_k \in \mathcal{D}}\mathbb{E}_{q_\phi(z|x_k)}\left[\frac{p(z)}{q_\phi(z)}\log{Z_k}\right]
\end{split} 
\end{equation*}
Adding the entropy term, the final result is:
\begin{equation*}
\begin{split}
    &\KLD{p_\theta(x)}{p_{\tau, \mathcal{D}}(x)} \le \\
    & \frac{1}{\tau|\mathcal{D}|}\sum_{x_k \in \mathcal{D}}\mathbb{E}_{q_\phi(z|x_k)}\left[\frac{p(z)}{q_\phi(z)}\mathbb{E}_{p_\theta(x|z)}\left[D_{\text{edit}}(x, x_k)\right]\right] + \\ 
    &\frac{1}{|\mathcal{D}|}\sum_{x_k \in \mathcal{D}}\mathbb{E}_{q_\phi(z|x_k)}\left[\frac{p(z)}{q_\phi(z)}\log q_\phi(z|x_k)\right] - \\ 
    &\mathbb{E}_{p(z)}\log q_\phi(z) +\\
    &\frac{1}{|\mathcal{D}|}\sum_{x_k \in \mathcal{D}}\mathbb{E}_{q_\phi(z|x_k)}\left[\frac{p(z)}{q_\phi(z)}\log{Z_k}\right]- H(p_\theta(x))
\end{split} 
\end{equation*}
\section{Hyperparameters}
\label{sec:hyp_app}
To find optimal hyperparameters setting for the LSTM language model we used random search \citep{BergstraB12}, where all tuned hyperparameters were smapled from the grid in Table \ref{table:hyper_grid}.
The determined best hyperparameters for PTB dataset can be found in Table \ref{table:hyper_grid}.

\begin{table}[h]
\caption{Best hyperparameters and random search grid for LSTM-LM on PTB dataset.} 
\centering
\setlength\tabcolsep{4pt}
\label{table:hyper_grid}
\scalebox{0.65}{
\begin{tabular}{ccc}
\toprule
\textbf{Hyperparameter}                & \textbf{Range}                              &\textbf{Best} \\ \midrule
optimizer                              & [Adam, AdaDelta, Sgd]                       & Adam \\
embedding dim.                         & [128, 256, 512]                             & 256 \\
hidden dim.                            & [128, 256, 512]                             & 256 \\
embedding dropout rate                 & [none, 0.1, 0.2, 0.3, 0.4, 0.5]             & 0.4 \\
output (word classifier) dropout rate  & [none, 0.1, 0.2, 0.3, 0.4, 0.5]             & 0.4 \\
recurrent dropout rate                 & [none, 0.1, 0.2, 0.3, 0.4, 0.5]             & none \\
recurrent layer normalization          & [present, absent]                           & present \\
tied embedding and output weights      & [tied, untied]                              & tied \\
learning rate                          & [\(1, 10^{-1}, 10^{-2}, 10^{-3}, 10^{-4}\)] & \(10^{-2}\) \\
\(L_2\) regularization weight          & [\(10^{-2}, 10^{-3}, 10^{-4}, 10^{-5}\)]    & \(10^{-4}\) \\
max gradient Chebyshev norm            & [1.0, 3.0, \(\inf\)]                        & 3.0 \\
\bottomrule
\end{tabular}
}
\end{table}

We used tied weight for embedding and word classification weights \citep{InanKS17}.
Gradient clipping \citep{PascanuMB13} was employed to make learning more stable.
Layer normalization \citep{BaKH16} and dropout \citep{SemeniutaSB16} were used in recurrent layers to reduce an overfitting issue.

We kept less sensitive hyperparameters to the fixed values, which can be found in Table \ref{table:hyper_fixed}.
\begin{table}[h]
\caption{Fixed hyperparameters for LSTM-LM on PTB dataset.} 
\centering
\label{table:hyper_fixed}
\scalebox{0.65}{
\begin{tabular}{cc}
\toprule
\textbf{Hyperparameter}  & \textbf{Value} \\ \midrule
batch size                                    & 64 \\
sentence max length                           & 100 words \\
validation frequency                          & 3 times per epoch\\
early stopping patience                       & 7 epochs \\
early stopping improvement threshold          & relative 0.1\% \\
max number of epochs                          & 100 epochs \\
learning rate scheduler patience              & 3 epochs \\
learning rate reducing multiplicative factor  & 0.5\\
\bottomrule
\end{tabular}
}
\end{table}

\begin{table}[h]
\caption{Hyperparameters for VAE models on PTB dataset.} 
\centering
\label{table:vae_hyper}
\scalebox{0.65}{
\begin{tabular}{ccc}
\toprule
\textbf{Hyperparameter}                          & \textbf{Range}  & \textbf{Value} \\ \midrule
latent code dim.                                      & --              & 32 \\
encoder embedding dropout rate                   & --              & 0.4 \\
encoder recurrent dropout rate                   & --              & 0.1 \\
encoder recurrent layer normalization            & --              & present \\
batch size                                       & --              & 256 \\
generated seq. max length (for Lev. VAE) & --                  & 120 \\
\(\beta\) (for \(\beta\)-VAE)                    & [0.1; 1.0; step=0.05]        & 0.6 \\
\(M\) (cycle period for Cyclical ann. VAE)  & [1; 10; step=1]              & 5 \\
\(\alpha\) (for Levenshtein VAE, see Eq.\ref{eq:final_gradients}) & -- & 1.0 \\
\(\lambda\) (for Levenshtein VAE, see Eq.\ref{eq:final_gradients}) & [0.5; 2.0; step=0.05] & 0.15 \\
\(\tau\) (for Levenshtein VAE, see Eq.\ref{eq:final_gradients}) & -- & 1.0\\
\bottomrule
\end{tabular}
}
\end{table}

\begin{table}[h]
\caption{Hyperparameters for VAE models on Yelp dataset.}
\centering
\label{table:yelp_vae_hyper}
\scalebox{0.65}{
\begin{tabular}{ccc}
\toprule
\textbf{Hyperparameter}                          & \textbf{Range}  & \textbf{Value} \\ \midrule
latent code dim.                                 & --              & 256 \\
hidden dim.                                      & --              & 1024 \\
embedding dim.                                   & --              & 512 \\
encoder embedding dropout rate                   & --              & 0.4 \\
encoder recurrent dropout rate                   & --              & 0.1 \\
encoder recurrent layer normalization            & --              & present \\
batch size                                       & --              & 128 \\
generated seq. max length (for Lev. VAE)         & --              & 30 \\
\(\beta\) (for \(\beta\)-VAE)                    & [0.0; 1.0; step=0.1]        & 1.0 \\
\(M\) (cycle period for Cyclical ann. VAE)  & [2; 18; step=2]              & 18 \\
\(\alpha\) (for Levenshtein VAE\(_{\alpha=0}\), see Eq.\ref{eq:final_gradients}) & -- & 0.0 \\
\(\lambda\) (for Levenshtein VAE\(_{\alpha=0}\), see Eq.\ref{eq:final_gradients}) & -- & 1.0 \\
\(\tau\) (for Levenshtein VAE\(_{\alpha=0}\) see Eq.\ref{eq:final_gradients}) & [0.0; 1.0; step=0.1] & 0.2 \\
\(\alpha\) (for Levenshtein VAE\(_{\alpha=1.0}\), see Eq.\ref{eq:final_gradients}) & -- & 1.0 \\
\(\lambda\) (for Levenshtein VAE\(_{\alpha=1.0}\), see Eq.\ref{eq:final_gradients}) & -- & 1.0 \\
\(\tau\) (for Levenshtein VAE\(_{\alpha=1.0}\), see Eq.\ref{eq:final_gradients}) & [0.1; 2.2; step=0.1] & 0.7\\
\bottomrule
\end{tabular}
}
\end{table}

\begin{table}[h]
\caption{Hyperparameters for VAE models on SNLI dataset.}
\centering
\label{table:snli_vae_hyper}
\scalebox{0.65}{
\begin{tabular}{ccc}
\toprule
\textbf{Hyperparameter}                          & \textbf{Range}  & \textbf{Value} \\ \midrule
latent code dim.                                 & --              & 256 \\
hidden dim.                                      & --              & 1024 \\
embedding dim.                                   & --              & 512 \\
encoder embedding dropout rate                   & --              & 0.4 \\
encoder recurrent dropout rate                   & --              & 0.1 \\
encoder recurrent layer normalization            & --              & present \\
generated seq. max length (for Lev. VAE)         & --              & 120 \\
\(\beta\) (for \(\beta\)-VAE)                    & [0.0; 1.0; step=0.1]        & 0.1 \\
\(M\) (cycle period for Cyclical ann. VAE)  & [2; 18; step=4]              & 2 \\
\(\alpha\) (for Levenshtein VAE\(_{\alpha=0}\), see Eq.\ref{eq:final_gradients}) & -- & 0.0 \\
\(\lambda\) (for Levenshtein VAE\(_{\alpha=0}\), see Eq.\ref{eq:final_gradients}) & -- & 1.0 \\
\(\tau\) (for Levenshtein VAE\(_{\alpha=0}\) see Eq.\ref{eq:final_gradients}) & [0.0; 1.0; step=0.1] & 0.4 \\
\(\alpha\) (for Levenshtein VAE\(_{\alpha=1.0}\), see Eq.\ref{eq:final_gradients}) & -- & 1.0 \\
\(\lambda\) (for Levenshtein VAE\(_{\alpha=1.0}\), see Eq.\ref{eq:final_gradients}) & -- & 1.0 \\
\(\tau\) (for Levenshtein VAE\(_{\alpha=1.0}\), see Eq.\ref{eq:final_gradients}) & [0.1; 2.2; step=0.1] & 0.6\\
\bottomrule
\end{tabular}
}
\end{table}

\end{document}